\documentclass[10pt,conference]{IEEEtran}
\IEEEoverridecommandlockouts
\usepackage{graphicx}
\usepackage{setspace}
\usepackage{amsmath}
\usepackage{dsfont}
\usepackage{amsfonts}
\usepackage{lipsum} 
\usepackage{multicol}
\usepackage{float}
\usepackage{bm}
\usepackage{color}
\usepackage{etoolbox}
\usepackage{soul}
\usepackage{amssymb}
\usepackage[linesnumbered,ruled,vlined]{algorithm2e}
\usepackage{mathtools,amssymb}
\usepackage{caption}
\usepackage{float}
\usepackage{amsthm}
\usepackage[caption = false,subrefformat=parens,labelformat=parens]{subfig}
\usepackage{adjustbox}
\usepackage{afterpage}
\usepackage{multirow}
\usepackage{mathrsfs}
\usepackage{booktabs}
\usepackage{url}
\usepackage{xcolor}
\definecolor{red}{HTML}{9E3E26}
\definecolor{blue}{HTML}{0B6B91}
\title{Embedding Byzantine Fault Tolerance into Federated Learning via Consistency Scoring

\author{
Youngjoon Lee\textsuperscript{\rm 1}, Jinu Gong\textsuperscript{\rm 2}, Joonhyuk Kang\textsuperscript{\rm 1} \\
\textsuperscript{\rm 1}School of Electrical Engineering, KAIST, South Korea\\
\textsuperscript{\rm 2}Department of Applied AI, Hansung University, South Korea\\
Email: yjlee22@kaist.ac.kr, jinugong@hansung.kr, jkang@kaist.ac.kr
}

\thanks{This research was partly supported by the Institute of Information \& Communications Technology Planning \& Evaluation (IITP)-ITRC (Information Technology Research Center) grant funded by the Korea government (MSIT) (IITP-2025-RS-2020-II201787, contribution rate: 50\%) and (IITP-2025-RS-2023-00259991, contribution rate: 50\%).
}
}

\begin{document}

\maketitle

\begin{abstract}
Given sufficient data from multiple edge devices, federated learning (FL) enables training a shared model without transmitting private data to the central server. 
However, FL is generally vulnerable to Byzantine attacks from compromised edge devices, which can significantly degrade the model performance.
In this work, we propose an intuitive plugin that seamlessly embeds Byzantine resilience into existing FL methods.
The key idea is to generate virtual data samples and evaluate model consistency scores across local updates to effectively filter out compromised updates.
By utilizing this scoring mechanism before the aggregation phase, the proposed plugin enables existing FL methods to become robust against Byzantine attacks while maintaining their original benefits.
Numerical results on blood cell classification task demonstrate that the proposed plugin provides strong Byzantine resilience. 
In detail, plugin-attached FedAvg achieves over 89.6\% test accuracy under 30\% targeted attacks (vs.\ 19.5\% w/o plugin) and maintains 65--70\% test accuracy under untargeted attacks (vs.\ 17--19\% w/o plugin).

\end{abstract}

\section{Introduction}
\label{sec:intro}
Recent advances in deep learning have transformed healthcare by leveraging large-scale medical datasets \cite{bisio2025ai, bisio2023feet}. 
However, strict patient privacy regulations and HIPAA compliance requirements have created significant barriers to traditional centralized analysis of healthcare data \cite{thapa2021precision}. 
Federated learning (FL) \cite{mcmahan2017communication} has emerged as a promising framework for healthcare applications \cite{lee2025revisit}. 
This method enables collaborative learning of medical AI models while preserving patient privacy through secure parameter sharing \cite{lee2023fast}.
For example, clinical diagnosis and medical imaging analysis have shown significant improvements using this privacy-preserving paradigm \cite{joshi2022federated}. 

\begin{figure}
     \centering
     \includegraphics[width=\columnwidth]{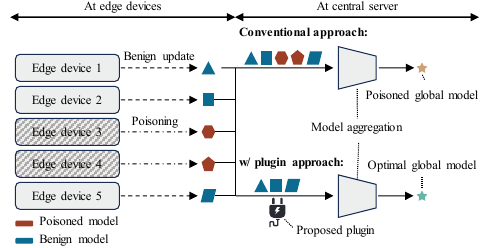}
     \caption{Comparison of conventional FL and plugin-enhanced FL under Byzantine attacks. In the conventional approach (top), malicious edge devices (devices 3 and 4) inject poisoned model updates that compromise the global model during aggregation. Our proposed plugin (bottom) effectively filters out these poisoned updates to maintain model integrity, leading to an optimal global model.}
     \label{fig:fig1}
\end{figure}

While FL offers privacy benefits, implementing it in real-world healthcare systems faces significant challenges \cite{li2020federated}. 
Healthcare data collected from different medical institutions naturally exhibits data heterogeneity due to varying patient populations and clinical protocols \cite{kairouz2021advances,9874855}. 
Moreover, Byzantine threats pose a critical concern in federated medical systems, as malicious participants can compromise the global model through Byzantine attacks with poisoned updates \cite{bagdasaryan2020backdoor}. 
This vulnerability is particularly concerning in clinical applications where model failures could directly impact patient safety and care outcomes \cite{nguyen2022federated}.
Therefore, developing Byzantine robust FL against adversarial attacks while maintaining model performance under heterogeneous conditions remains an important challenge.

The novel plugin-based architecture proposed in Fig. \ref{fig:fig1} embeds Byzantine resilience into existing FL methods without compromising their original benefits. 
Our plugin generates virtual data samples to evaluate model behavior patterns, computing consistency scores across local updates to detect and filter malicious contributions while preserving the core functionality of FL methods. 
Extensive experiments on medical imaging datasets demonstrate that this approach enhances robustness while maintaining high performance across diverse FL methods.
Main contributions of this paper are as follows:
\begin{itemize}
    \item We propose a novel plugin to embed Byzantine resilience into FL methods without altering their core principle.
    \item We propose a virtual data-driven scoring method to detect and filter compromised local updates.
    \item We show our plugin's compatibility and validate its effectiveness through comprehensive experiments.
\end{itemize}

The remainder of this paper is organized as follows. 
Section \ref{sec:preliminaries} provides background on representative FL methods. 
Section \ref{sec:main} presents our novel modular plugin for FL.
Section \ref{sec:experiment} demonstrate how our plugin enhances performance against Byzantine attacks through extensive numerical evaluations.

\section{Preliminaries}
\label{sec:preliminaries}
To describe FL methods, we consider a federated network with $K$ edge devices, where each device $k$ holds a local dataset $\mathcal{D}_k$. 
The goal of FL is to solve:
\begin{equation}
\min_{w} F(w) = \sum_{k=1}^K F_k(w),
\end{equation}
where $F_k(w) = \frac{1}{|\mathcal{D}_k|}\sum_{(x,y)\in\mathcal{D}_k}\ell(w;x,y)$ represents the local objective function of edge device $k$.

\textbf{FedProx} \cite{li2018federated} addresses data heterogeneity by introducing a proximal term in the local objective function, enforcing local model updates to remain close to the global model:
\begin{equation}
F_k^{Prox}(w) = F_k(w) + \frac{\mu}{2}|w - w^{g_e}|^2,
\end{equation}
where $g_e$ denotes global epoch and $\mu$ controls the client drift. 
This proximal regularization allows FedProx to achieve more stable convergence even when data across devices is highly heterogeneous.

\textbf{FedDyn} \cite{acarfederated} utilizes dynamic regularization to align local optimization objectives with the global goal by adapting to local optimization paths. The local objective in FedDyn is as follows:
\begin{equation}
F_k^{Dyn}(w) = F_k(w) + (h^{g_e})^T(w - w^{g_e}) + \frac{\alpha}{2}|w - w^{g_e}|^2,
\end{equation}
where $\alpha$ is a scaling factor and $h^{g_e}$ captures optimization trajectory differences.
Then, the central server aggregates local updates and adjusts the global model using dynamic control variates.
This adaptive approach allows the model to dynamically update regularization, ensuring a better fit to each device's unique data distribution.

\textbf{FedRS} \cite{li2021edrs} introduces a restricted softmax method to tackle label distribution heterogeneity as follows:
\begin{equation}
\psi_{i,c}^k = \frac{\exp(\alpha_c^k (w_c^k)^T h_i^k)}{\sum_{j=1}^C \exp(\alpha_j^k (w_j^k)^T h_i^k)},
\end{equation}
where $h_i^k$ is the feature vector of the $i$-th sample and $w_c^k$ is the classifier weight for class $c$ on edge device $k$, $\alpha_c^k$ is a scaling factor to limit updates of missing classes while maintaining normal softmax behavior for observed classes.

\textbf{FedSAM} \cite{qu2022generalized} incorporates Sharpness-Aware Minimization (SAM) \cite{foret2021sharpnessaware} to enhance model generalization on heterogeneous data. SAM modifies the objective to minimize sharp local optima by perturbing gradients:
\begin{equation}
w_k^{SAM} = w_k^{g_e,l_e} - \rho \frac{\nabla F_k(w_k^{g_e, l_e} + \epsilon)}{|\nabla F_k(w_k^{g_e, l_e} + \epsilon)|},
\end{equation}
where $l_e$ denotes local epoch and $\epsilon = \rho \frac{\nabla F_k(w_k^{g_e, l_e})}{|\nabla F_k(w_k^{g_e, l_e})|}$ is a perturbation that stabilizes convergence by directing models toward flatter minima.

\textbf{FedSpeed} \cite{sunfedspeed} accelerates convergence through prox-correction and gradient perturbation:
\begin{equation}
w^{g_e, l_e+1}_{k} = w^{g_e, l_e}_{k} - \eta(\tilde{\nabla F}^{g_e, l_e}_k - \hat{\nabla F}^{g_e}_k + \frac{1}{\lambda}(w^{g_e, l_e}_{k} - w^{g_e})),
\end{equation}
where $\eta$ denotes local learning rate and  $\tilde{\nabla F}^{g_e, l_e}_k$ is a quasi-gradient, and $\hat{\nabla F}^{g_e}_k$ is a prox-correction term. This technique mitigates client drift while maintaining high generalization through flat minima search, leading to faster convergence with larger local intervals.

\section{Problem and Model}\label{sec:main}
\subsection{Byzantine Attack and Non-IID Setting}
We consider an FL environment vulnerable to both targeted and untargeted model poisoning attacks in medical imaging task as \cite{lee2024security}.
Specifically, the FL system comprises $K$ participating edge devices, including $B$ benign and $M$ compromised nodes, all connected to a central server. 
Each edge device $k$ maintains its private patient dataset $\mathcal{D}_{k}$ with varying sizes. 
To preserve patient privacy, edge devices collaboratively train a shared medical image classifier over $G$ global epochs by sharing only model parameters with the central server.

\subsection{Plugin-based Byzantine-Resilient FL}
In this section, we introduce a plugin-based approach that enhances FedAvg's resilience against Byzantine attacks through feature-space analysis. 
First, benign edge devices perform local learning, while malicious edge devices craft Byzantine attacks.
For benign edge devices $b \in B$, local training aims to minimize the empirical loss:
\begin{equation}
F_b(w) = \frac{1}{|\mathcal{D}_b|} \sum_{(x,y) \in \mathcal{D}_b} \ell(w; x, y),
\end{equation}
where $\ell(\cdot)$ is the cross-entropy loss function.
The local updates are performed via SGD with learning rate $\eta$:
\begin{equation}
w_b^{g_e, l_e+1} = w_b^{g_e, l_e} - \eta \nabla F_b(w_b^{g_e, l_e}; \mathcal{B}_b),
\end{equation}
where $\mathcal{B}_b$ is a randomly sampled mini-batch from $\mathcal{D}_b$.

However, compromised edge devices $m \in M$ may perform malicious updates through either targeted or untargeted attacks.
In targeted attacks, compromised devices alter their local updates after $L$ local training epochs, to mislead specific samples while preserving overall performance.
\begin{align}
w_m^{g_e, L} &\underset{attack}{\leftarrow} w_m^{g_e, L} + \delta_m,\\
\delta_m &= \lambda(w_m^{g_e, L}-w^{g_e}),
\end{align}
where $\lambda$ is a boosting factor designed to amplify the attack's impact. 
For untargeted attacks, malicious devices inject arbitrary noise to degrade overall performance:
\begin{align}
w_m^{g_e} &\underset{attack}{\leftarrow} \tau(w^\prime-w^{g_e}),\\
w^\prime &\sim \mathcal{N}(\boldsymbol{0}, I),
\end{align}
with scaling factor $\tau$ and standard Gaussian noise.

\begin{figure}[t]
\centering
\includegraphics[width=\columnwidth]{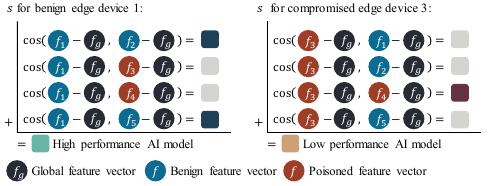}
\caption{Illustration of the proposed plugin's scoring principle. 
The plugin measures pairwise cosine similarities (denoted as `cos') between feature vectors of model updates ($f_1$ through $f_5$) relative to the global model ($f_g$). 
Benign models (\textcolor{blue}{\textbf{blue}}) have high cosine similarity with each other, whereas poisoned models (\textcolor{red}{\textbf{red}}) show distinct patterns, enabling effective Byzantine attack filtering.
}
\label{fig:fig2}
\end{figure}

To filter these attacks, our plugin performs Byzantine attack filtering using feature space analysis in three main steps: deviation analysis, feature mapping, and similarity-based rejection.
First, the central server generate $N$ virtual samples $\{v_n\}_{n=1}^N$ from a standard normal distribution $\mathcal{N}(\boldsymbol{0},I)$ to serve as probe points for analyzing model behaviors. 
For each model pair $(w_i^{g_e}, w_j^{g_e})$, the server compute their deviations from the global model as:
\begin{align}
\Delta w_i = w_i^{g_e} - w^{g_e}, \; \Delta w_j = w_j^{g_e} - w^{g_e}.
\end{align}
These deviation vectors capture how each local model update differs from the current global model.

Next, the central server maps these deviations to feature representations using a feature extractor $g_{\phi}(\cdot)$, which is implemented as all layers of the base model except the final classification layer. For each model pair $(w_i, w_j)$ and virtual sample $v_n$, the server computes:
\begin{align}
\mathcal{F}_i &= \{f_i^1, f_i^2, \dots, f_i^N\} = g_{1:L-1}(v_n; \Delta w_i), \\
\mathcal{F}_j &= \{f_j^1, f_j^2, \dots, f_j^N\} = g_{1:L-1}(v_n; \Delta w_j),
\end{align}
where $g_{1:L-1}$ denotes the feature extraction layers of the model (all layers except the final classification layer), and each $f_i^n, f_j^n \in \mathbb{R}^d$ represents the $d$-dimensional feature vector for the $n$-th virtual sample.

The server then computes the pairwise similarity between models using the average cosine similarity over the $N$ virtual samples:
\begin{equation}
s_{i,j} = \frac{1}{N}\sum_{n=1}^N \cos\big(f_i^n, f_j^n\big),
\ f_i^n \in \mathcal{F}_i,~ f_j^n \in \mathcal{F}_j.
\end{equation}
where $\cos(f_i^n, f_j^n)$ measures the angular similarity between feature vectors, with higher values indicating more similar behavior patterns.
As shown in Fig.~\ref{fig:fig2}, benign models naturally cluster together with high similarity scores, whereas malicious model updates exhibit distinct patterns. 
For each model update $w_k$, the server calculates its average cosine similarity with the other updates as:
\begin{equation}
\bar{s}_k = \frac{1}{K-1}\sum_{j \neq k} s_{k,j}.
\end{equation}

The server then sorts these average similarity scores $\{\bar{s}_k\}_{k=1}^K$ in ascending order and defines the set $\mathcal{S}$ as:
\begin{equation}
\mathcal{S} = \{k \mid \pi(\bar{s}_k) > M\},
\end{equation}
where $\pi(\bar{s}_k)$ denotes the position of $\bar{s}_k$ in the sorted scores, effectively excluding the $M$ model updates with the lowest similarity scores as potential Byzantine attacks. The Byzantine-resilient global model is then updated by averaging the remaining model updates:
\begin{equation}
w^{g_e+1} = \frac{1}{|\mathcal{S}|}\sum_{k \in \mathcal{S}} w_k.
\end{equation}
Finally, the central server broadcasts the aggregated model to all edge devices for the next global training round. 
The overall procedure of the proposed plugin for FL method is described in Algorithm 1.

\begin{algorithm}[h]
\caption{Proposed Modular Plugin for FL}
\SetAlgoLined
\KwIn{Local models $\{w_k^{g_e}\}_{k=1}^K$, global model $w^{g_e}$, number of malicious devices $M$}
\KwOut{Global model $w^{g_e+1}$}
\tcc{Virtual data-driven consistency scoring at the central server}
Generate $\{v_n\}_{n=1}^N \sim \mathcal{N}(\boldsymbol{0},I)$\;

\For{$i,j \in [K]$}{
    $\mathcal{F}_i \gets \mathbf{g}_{1:L-1}(v_n; \Delta w_i)$\;
    $\mathcal{F}_j \gets \mathbf{g}_{1:L-1}(v_n; \Delta w_j)$\;
    $s_{i,j} \gets \frac{1}{N}\sum_{n=1}^N \cos(f_i^n, f_j^n)$\;
}

$\bar{s}_k \gets \frac{1}{K-1}\sum_{j\neq k} s_{k,j}$ for all $k$\;
$\mathcal{S} = \{k \mid \pi(\bar{s}_k) > M\}$\;
Aggregate $w^{g_e+1} = \frac{1}{|\mathcal{S}|}\sum_{k\in\mathcal{S}} w_k$\;
\Return{$w^{g_e+1}$}
\end{algorithm}

\section{Experiment and Results}\label{sec:experiment}
\subsection{Experiment setting}
We evaluate our proposed plugin on blood cell classification task \cite{acevedo2020dataset}, where each edge device runs a ResNet-18 \cite{he2016deep} model as its local model.
Our experiments compare the performance of various FL methods with and without our proposed plugin under both Targeted Model Poisoning (TMP) and Untargeted Model Poisoning (UMP) attack scenarios.
To simulate realistic data heterogeneity, we distribute the data non-uniformly across $K=10$ medical edge devices following the quantity skew protocol \cite{li2022federated}.
The complete implementation details and configuration parameters are publicly available in our open repository\footnote{https://github.com/yjlee22/fl-plugin}.

\subsection{Results}

\subsubsection{Impact of Byzantine Attacks}
\begin{figure}[t]
   \centering
   
       \centering
       \subfloat[Targeted Model Poisoning Attack]
           {
           \includegraphics[width=\columnwidth]{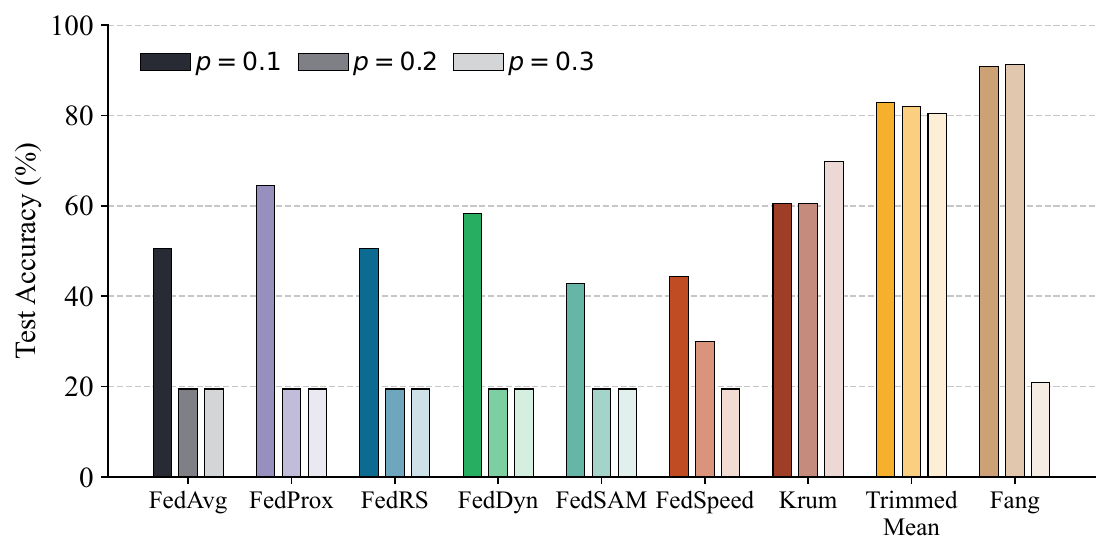}
           \label{fig:targeted}
           }

       \centering
       \subfloat[Untargeted Model Poisoning Attack]
           {
           \includegraphics[width=\columnwidth]{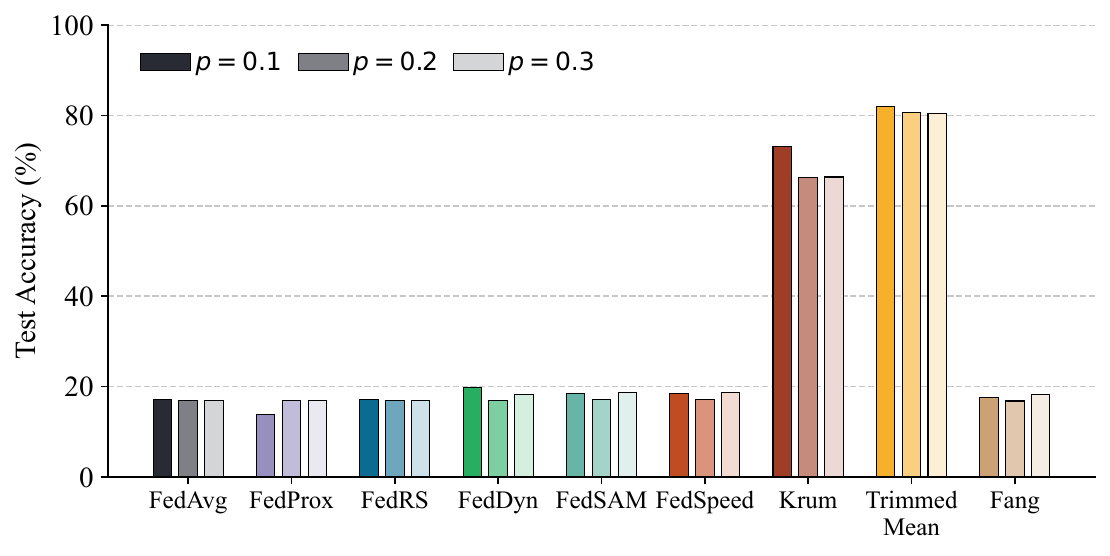}
           \label{fig:untargeted}
           }

   \caption{Performance comparison of heterogeneity-aware and Byzantine-resilient FL methods under model poisoning attacks. The x-axis denotes the ratio of compromised devices ($p$), and the y-axis shows the test accuracy (\%).}
   \label{fig:plot1}
\end{figure}

To examine the vulnerability of heterogeneity-aware FL methods, we check the performance under model poisoning attacks. 
As shown in Fig.~\ref{fig:plot1}, all vanilla FL methods experience severe degradation under Byzantine attack settings. 
As the proportion of compromised devices increases, their accuracy falls sharply, reaching about 20\% when $p=0.3$. 
This trend highlights the absence of mechanisms to identify and remove malicious updates. 
Moreover, untargeted attacks result in greater harm than targeted attacks even when the compromise ratio is low.
In detail, all heterogeneity-aware FL methods achieve less than 20\% accuracy when subjected to untargeted attacks.
Thus, additional Byzantine resilient components are essential for practical FL deployments.

In addition, we compare with representative Byzantine-resilient FL methods, including Krum~\cite{blanchard2017machine}, Trimmed-Mean~\cite{yin2018byzantine}, and Fang~\cite{fang2020local}. 
These Byzantine-resilient FL methods consistently outperform heterogeneity-aware ones in adversarial settings.
The noticeable performance gap underscores the severity of Byzantine threats in realistic heterogeneous networks. 
Furthermore, untargeted attacks remain highly disruptive due to their random and unpredictable perturbations. 
Hence, secure FL requires mechanisms beyond addressing data heterogeneity to counter adaptive attackers. 

\subsubsection{Impact of Different Compromise Fractions}
To check the effect of the proposed plugin, we evaluate its performance under varying fractions of compromised devices. 
As shown in the top of Fig.~\ref{fig:plot2}, our plugin shows remarkable effectiveness across all FL methods under targeted attack.
Specifically, vanilla FedAvg accuracy drops from 80.06\% to 19.47\% as $p$ rises to 0.3, reflecting extreme vulnerability. 
However, with the plugin enabled, FedAvg maintains above 89\% even at the highest compromise level by filtering not only Byzantine updates but also unhelpful local updates. 
Similar improvements appear in FedProx, FedDyn, and FedRS, which sustain 85--90\% accuracy across all values of $p$.
The results indicate that the proposed plugin’s ability to detect and filter malicious updates under targeted attacks. 

However, as shown in the bottom of Fig.~\ref{fig:plot2}, training under untargeted attacks remains more challenging due to their random nature. 
All vanilla FL methods collapse below 20\% accuracy even when $p=0.1$, highlighting their vulnerability. 
In contrast, plugin-attached FedAvg, FedProx, FedDyn, and FedRS achieve 65--70\% accuracy at $p=0.3$, showing considerable gains. 
As $p$ increases, the gap between vanilla and plugin-attached versions widens markedly, emphasizing the plugin’s growing importance. 
Thus, our plugin provides meaningful filtering under both attack types and is most effective against targeted attacks.

\subsubsection{Impact of Plugin with Different Non-IID Degrees}
Finally, we analyze how varying data heterogeneity affects the plugin’s filtering capabilities under Byzantine attacks with $p=0.3$. 
As shown in the top of Fig.~\ref{fig:plot3}, all vanilla FL methods remain stuck near 20\% accuracy regardless of $\log\alpha$.
However, plugin-attached FedAvg, FedProx, FedDyn, and FedRS achieve 85--90\% at $\log\alpha=1$ under targeted attacks, demonstrating substantial gains. 
FedSAM and FedSpeed follow similar trends, reaching 70--80\% under the same conditions.
Hence, plugin effectiveness improves as data distributions become more homogeneous and stable. 
This relationship highlights the interplay between heterogeneity levels and Byzantine resilience degree.

Under untargeted attacks, as shown in the bottom of Fig.~\ref{fig:plot3}, the impact of heterogeneity is even more pronounced across all FL methods. 
In particular, vanilla FL methods remain weak around 17--20\% accuracy for all $\alpha$ values, confirming their inherent limitations. 
Meanwhile, plugin-attached FL methods steadily improve with increasing $\log\alpha$, showing adaptability to distributional shifts. 
At $\log\alpha=2$, FedAvg, FedProx, FedDyn, and FedRS reach 74--75\%, while FedSAM and FedSpeed attain 64--65\%. 
However, when heterogeneity is high ($\log\alpha<0$), even FL methods with plugin struggle to exceed 30\%, indicating persistent challenges. 
The performance gap between TMP and UMP scenarios remains about 15--20\% at high $\alpha$, reflecting untargeted attack complexity. 
However, consistent improvement as $\alpha$ grows confirms that our plugin adapts well to heterogeneous FL settings.

\afterpage{
\newpage
\begin{figure*}
\centering
\includegraphics[width=\textwidth]{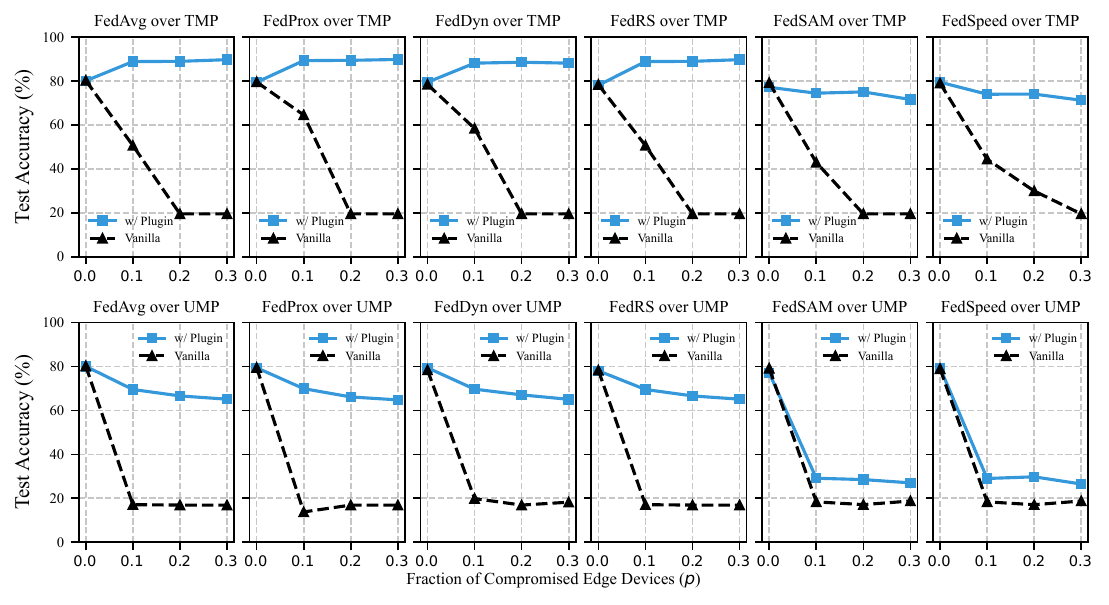}
   \caption{Impact of the proposed plugin on FL methods under TMP and UMP. For each method, we compare the test accuracy between vanilla and plugin-attached versions across different fractions of compromised devices ($p$).}
\label{fig:plot2}
\vspace{-0.2cm}
\end{figure*}

\begin{figure*}
\centering
\includegraphics[width=\textwidth]{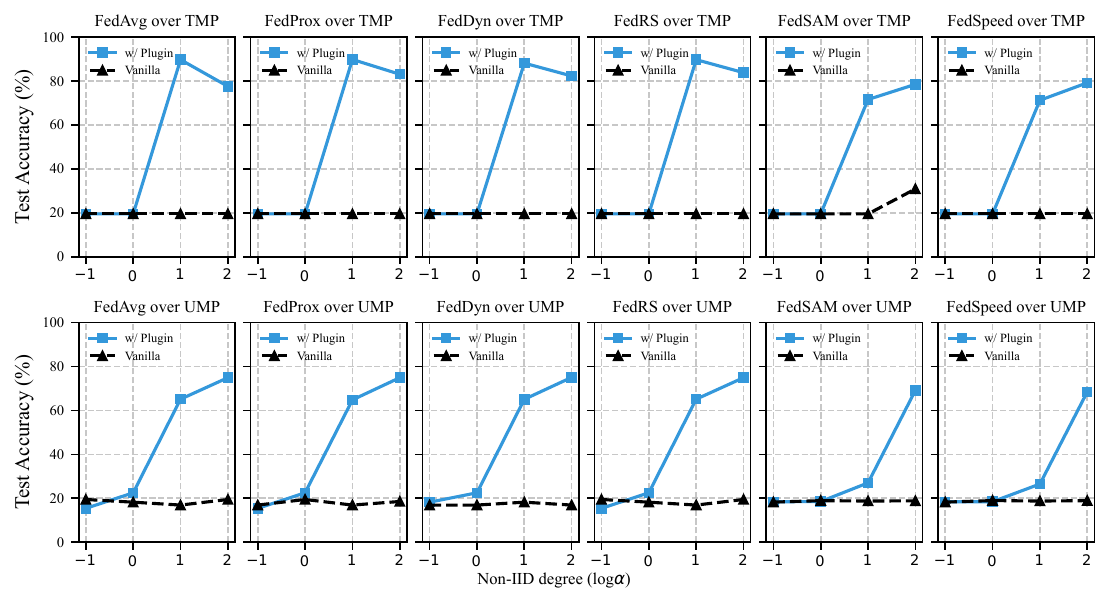}
   \caption{Impact of the proposed plugin on FL methods under TMP and UMP with $p=0.3$. For each method, we compare the test accuracy between vanilla and plugin-attached versions across different Non-IID degree ($\alpha$).}
\label{fig:plot3}
\vspace{-0.17cm}
\end{figure*}
\clearpage
}

\setlength{\columnsep}{0.21in}

\section{Conclusion}\label{sec:conclusion}
In this work, we propose a intuitive plugin-based approach to embed Byzantine resilience into heterogeneity-aware FL methods. 
Moreover, our solution can be attached seamlessly without modifying the core of FL methods.
Through extensive experiments on blood cell classification task, we validate the effectiveness of the proposed plugin. 
In particular, the virtual data-driven consistency scoring accurately detects and filters malicious updates.
Thus, our plugin offers a practical way to enable FL methods robust to Byzantine attacks.

\bibliographystyle{IEEEtran}
\bibliography{reference}

\end{document}